\documentclass[runningheads]{llncs}
\usepackage{graphicx}
\usepackage{amssymb, amsmath, bm, latexsym,comment}
\usepackage{multirow}
\usepackage{xcolor}
\usepackage{subfigure}
\usepackage{indentfirst}
\usepackage{setspace}
\usepackage{verbatim}
\usepackage{array}
\usepackage{arydshln}
\usepackage[misc]{ifsym} 
\usepackage[colorlinks,linkcolor=blue]{hyperref}
\usepackage{xcolor}
\usepackage[normalem]{ulem} 

\providecommand{\Leireffig}[1]{Fig.~\ref{#1}}
\providecommand{\Leireffigure}[1]{Figure~\ref{#1}}

\providecommand{\citep}[1]{\cite{#1}}

\newcommand{\blue}{\textcolor{black}}

\begin{document}
\title{Influence of Myocardial Infarction on QRS Properties: A Simulation Study}

\author{ 
Lei Li \inst{1\dagger} \and
Julia Camps \inst{2\dagger} \and 
Zhinuo (Jenny) Wang \inst{2} \and 
Abhirup Banerjee \inst{1,3}\orcidID{0000-0001-8198-5128} \and 
Blanca Rodriguez \inst{2} \and 
Vicente Grau \inst{1}\orcidID{0000-0001-8139-3480} 
} 
\authorrunning{Lei Li et~al.}

\institute{
Department of Engineering Science, University of Oxford, Oxford, UK \and
Department of Computer Science, University of Oxford, Oxford, UK \and
Division of Cardiovascular Medicine, Radcliffe Department of Medicine, University of Oxford, Oxford, UK \\
\email{$^{\dagger}$Two authors contribute equally. \\ lei.li@eng.ox.ac.uk;julia.camps@cs.ox.ac.uk}
}

\maketitle 


\begin{abstract}
The interplay between structural and electrical changes in the heart after myocardial infarction (MI) plays a key role in the initiation and maintenance of arrhythmia. 
The anatomical and electrophysiological properties of scar, border zone, and normal myocardium modify the electrocardiographic morphology, which is routinely analysed in clinical settings. 
However, the influence of various MI properties on the QRS is not intuitively predictable.
In this work, we have systematically investigated the effects of 17 post-MI scenarios, varying the location, size, transmural extent, and conductive level of scarring and border zone area, on the forward-calculated QRS. 
Additionally, we have compared the contributions of different QRS score criteria for quantifying post-MI pathophysiology.
The propagation of electrical activity in the ventricles is simulated via a Eikonal model on a unified coordinate system.
The analysis has been performed on 49 subjects, and the results imply that the QRS is capable of identifying MI, suggesting the feasibility of inversely reconstructing infarct regions from QRS.
There exist sensitivity variations of different QRS criteria for identifying 17 MI scenarios, which is informative for solving the inverse problem.

\end{abstract}

\section{Introduction}

Myocardial infarction (MI) is a major cause of mortality and disability worldwide \citep{journal/lancet/john2012,journal/EHJ/thygesen2019}.
Assessment of myocardial viability is essential in the diagnosis and treatment management for patients suffering from MI.
In particular, the position, size, and shape of the scarring region and the border zone could provide important information for the selection of patients and delivery of therapies for MI.
The electrocardiogram (ECG) is one of the most commonly used clinical diagnostic tools for MI \citep{journal/NEJM/zimetbaum2003}.
It can provide useful information about the heart rhythm and reveal abnormalities related to the conduction system \citep{conf/STACOM/li2023}.
For example, ST-segment elevation and T-wave inversion are widely investigated indicators of cardiac remodeling associated with different stages of MI \citep{journal/JE/hanna2011}.
In contrast, the QRS patterns has received less attention when analyzing ECG abnormalities associated with MI.
It is not yet fully clear how QRS abnormalities reflect MI characteristics, with some previous papers reporting conflicting results \citep{journal/JE/strauss2009,journal/EP/wang2021}.

\textit{In-silico} computer ECG simulations offer a powerful tool for mechanistic investigations on the MI characteristics \citep{journal/JCP/neic2017,journal/EP/wang2021}.
For example, Arevalo et~al. constructed a cardiac computational model, where simulations of the electrical activity were executed for arrhythmia risk stratification of MI patients \citep{journal/NC/arevalo2016}.
Wang et~al. developed a multi-scale cardiac modeling and ECG simulation framework for mechanistic investigations into the pathophysiological ECG and mechanical behavior post-MI \citep{journal/EP/wang2021}.
Costa et~al. employed a computational ventricular model of porcine MI to investigate the impact of model anatomy, MI morphology, and EP personalization strategies on simulated ECGs.
Que et~al. designed a multi-scale heart-torso computational model to simulate pathological 12-lead post-MI ECGs with various topographies and extents for ECG data augmentation \citep{journal/CMPB/que2022}.

In this work, we investigate the association between QRS abnormalities and MI characteristics in a unified coordinate system.
In this preliminary study, we only investigate QRS morphology rather than the complete ECG cycle, as the QRS simulation is quite efficient compared to the whole cycle.
For each subject, we examine 17 MI scenarios, summarize their effects on the simulated QRS, and identify the scenarios with the most significant alterations in the QRS morphology.
This study highlights the potential of QRS to improve the identification and localization of MI and further facilitate patient-specific clinical decision-making. 
It also demonstrates the feasibility of developing a cardiac ``digital twin" deep computational model for the inference of MI by solving an inverse problem.
The computational model provides an integrated perspective for each individual that incorporates the features from multi-modality data on cardiac systems.
To the best of our knowledge, this is the first sensitivity analysis of QRS complex for quantifying the MI characteristic variation in cardiac electrical activities.

\section{Methodology}

\subsection{Anatomical Model Construction} \label{method:cobiveco}

\begin{figure*}[t]\center
 \includegraphics[width=0.9\textwidth]{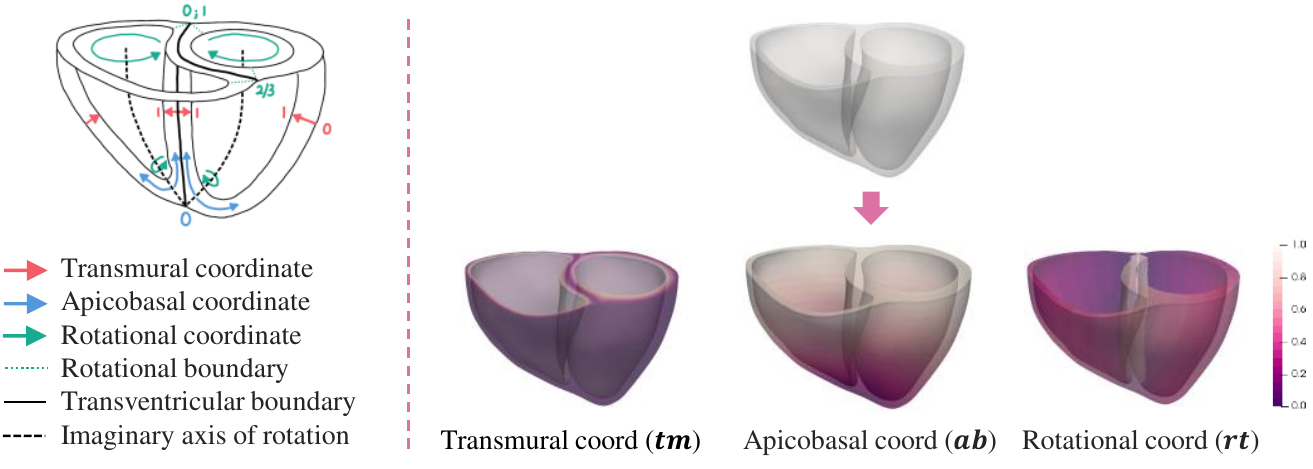}\\[-2ex]
   \caption{Consistent biventricular coordinates in the cobiveco system.}
\label{fig:method:cobiveco}
\end{figure*}

To obtain anatomical information, we generate a subject-specific 3D biventricular tetrahedral mesh from multi-view cardiac magnetic resonance (CMR) images for each subject, using the method outlined in \citep{journal/PTRSA/banerjee2021}. 
We employ the cobiveco coordinate reference system for mesh representation to ensure a symmetric, consistent and intuitive biventricular coordinate system across various geometries \citep{journal/MedIA/schuler2021}.
The cobiveco coordinate is represented by $(tm, ab, rt)$, where $tm$, $ab$, and $rt$ refer to transmural, apicobasal, and rotational coordinates, respectively.
\Leireffigure{fig:method:cobiveco} presents the cobiveco coordinate system.

We use ellipses with radii ${tm}_r$, ${ab}_r$,  and ${rt}_r$ to represent infarct regions in the myocardium, represented as
\begin{equation}
	\frac {({tm}_i - {tm}_0)^2} {{{tm}_r}^2} + \frac {({ab}_i - {ab}_0)^2} {{{ab}_r}^2} + \frac {({rt}_i - {rt}_0)^2} {{{rt}_r}^2} \leq 1,
\end{equation}
where $({tm}_0, {ab}_0, {rt}_0)$ is the center coordinate of the infarct region.
To study the effects of MI location at a population level, we employ the American Heart Association (AHA) 17-segment model \citep{journal/EHJCI/lang2015} and consistently select the infarct areas and the transmural extent via cobiveco.

\subsection{Electrophysiological Simulation} \label{method:simulation}
Cardiac electrophysiology is simulated via an efficient orthotropic Eikonal model \citep{journal/MedIA/camps2021,journal/TBME/wallman2012} that incorporates a human-based Purkinje system into the formulation of the root node (RN) activation times.
The simulation is performed over the generated cobiveco mesh in Sec.~\ref{method:cobiveco} and can be defined as,
\begin{equation}\label{eq:eikonal_eq}
\left\{
\begin{split}
& \sqrt{\nabla^T t\mathcal{V}^2 \nabla t} = 1, \\
& t(\Gamma_0) = pk(\Gamma_0)-\min(pk(\Gamma_0)),
\end{split}
\right.
\end{equation}
where $\mathcal{V}$ are the orthogonal conduction velocities (CVs) of fibre, sheet (transmural), and sheet-normal directions, $t$ is the time at which the activation wavefront reaches each point in the mesh, $\Gamma_0$ is the set of locations (\emph{i.e.,} RNs) in the endocardium, and $pk$ is a Purkinje-tree delay function from the His-bundle to every point in the mesh. 
Thus, the earliest activation time at the RNs is defined as their delay from the His-bundle through the Purkinje tree normalized by the earliest activation.
The QRS can be calculated from the activation time map via a pseudo-ECG equation \citep{journal/CR/gima2002} for a 1D cable source with constant conductivity at a given electrode location $(x',y',z')$, as
\begin{equation}
\phi_e (x',y',z' ) = \frac{a^2 \sigma_i}{4 \sigma_e} \int - \nabla V_m \cdot \Big [ \nabla \frac{1}{r} \Big ] \,dx \,dy \,dz \ ,
\end{equation}
where $V_m$ is the transmembrane potential, $\nabla V_m$ is its spatial gradient, $r$ is the Euclidean distance from a given point $(x,y,z)$ to the electrode location, $a$ is a constant that depends on the fiber radius, and $\sigma_i$ and $\sigma_e$ are the intracellular and extracellular conductivities, respectively. 
QRS is obtained by considering this integral throughout the ventricular activation sequence period.
\blue{For the measurement of electrode locations, we utilize the automated 3D torso reconstruction pipeline from the CMR images \citep{conf/EMBC/smith2022}.
Note that the pseudo-ECG method can efficiently produce normalized ECG signals with a comparable level of morphological information as the bidomain simulation \citep{journal/FiP/minchole2019}.}

\begin{figure*}[t]\center
 \includegraphics[width=0.99\textwidth]{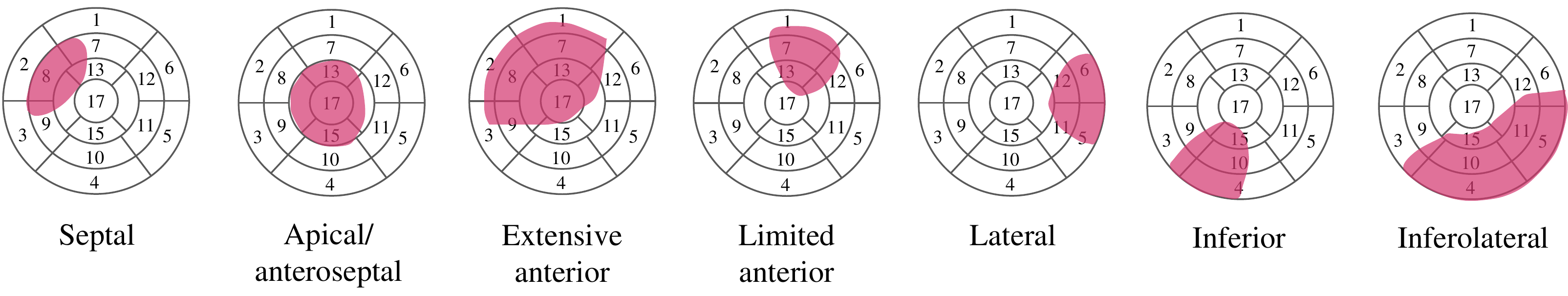}\\[-2ex]
   \caption{Seven MI locations on 17-segment AHA-map.}
\label{fig:method:MI_location}
\end{figure*}

\begin{table*} [t] \center
    \caption{
    Summary of the investigated MI scenarios. Ext: extensive; Lim: limited. 
     }
\label{tb:method:MI scenario}
{\scriptsize	
\begin{tabular}{p{2.4cm}|p{1cm}p{1cm}p{1.6cm}p{1.6cm}p{1.1cm}p{1.1cm}p{1.5cm}} 
\hline
Location           & Septal & Apical & Ext anterior & Lim anterior & Lateral & Inferior & Inferolateral \\
\hline
Transmural extent  & \checkmark & \checkmark  & \checkmark & \checkmark & \checkmark & \checkmark & \checkmark \\
Size               &            &             &            &            & \checkmark &            &            \\
CV in MI region    &            &             &            &            & \checkmark &            &            \\
\hline
\end{tabular} }\\
\end{table*}

\begin{figure*}[t]\center
 \includegraphics[width=0.99\textwidth]{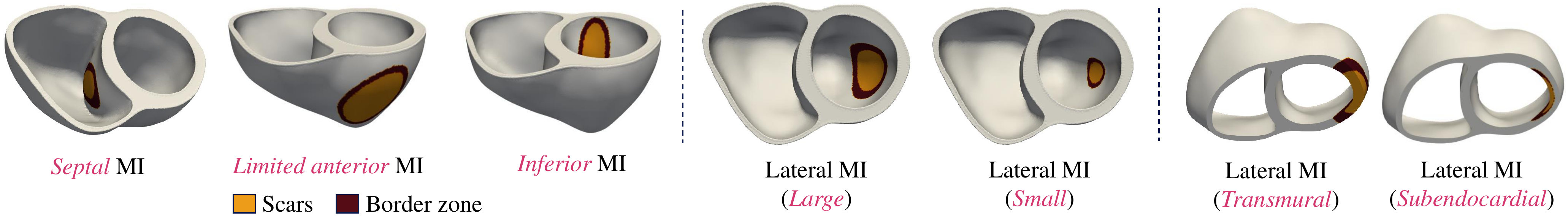}\\[-2ex]
   \caption{Illustration of several MI scenarios, including different MI locations, sizes, and transmural extents.}
\label{fig:method:MI_examples}
\end{figure*}

For simulation, we consider electrophysiological heterogeneities in the infarct regions, including \emph{seven locations (see \Leireffig{fig:method:MI_location}), two transmural extents (transmural and subendocardial MI), two different sizes, and two different sets of slower CVs in the infarct areas} \citep{journal/SR/martinez2019}.
Note that for the comparison of different MI sizes and CV decreasing extents, we only report on lateral MI as an illustrative case.
Therefore, for each subject we simulate 17 heterogeneous MI scenarios and one normal ECG as the baseline.
\Leireffigure{fig:method:MI_examples} provides examples of generated MI heterogeneity scenarios. 
We vary the CVs of infarct and healthy myocardial areas during its simulation, as slower CVs have been observed in the infarcted human heart \citep{journal/Circulation/de1993}.
Conduction pathways for electrical propagation in the infarct regions might exist, as observed in clinical data \citep{journal/CAE/strauss2008}. 
Therefore, we set the CVs of scarring and border zone areas to 10\% and 50\% (another CV set: 5\% and 25\%) of the values in healthy myocardium, respectively.

\subsection{Univariate Sensitivity Analysis} \label{method:SenAna}

\begin{figure*}[t]\center
 \includegraphics[width=0.9\textwidth]{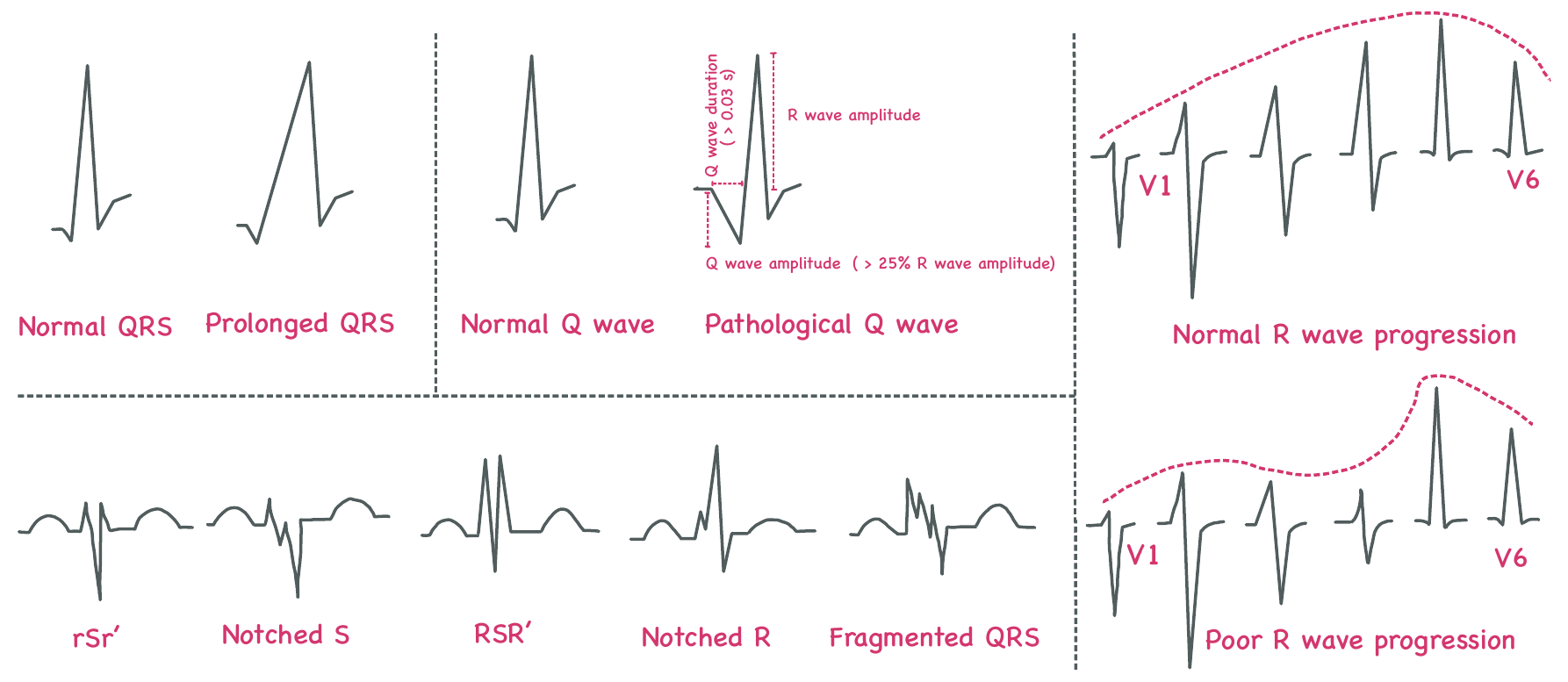}\\[-2ex]
   \caption{Sketch map of normal QRS and MI-related QRS abnormalities.} 
\label{fig:method:abnormalQRS_MI_example}
\end{figure*}

In the sensitivity analysis, we introduce a global QRS measure, dynamic time warping (DTW), to calculate the dissimilarity of QRS with different lengths \citep{journal/MedIA/camps2021}.
Moreover, we investigate four local QRS criteria, corresponding to QRS abnormalities of MI reported in the literature, namely, \emph{QRS duration prolongation} \citep{journal/CJ/cupa2018}, \emph{pathological Q-waves} \citep{journal/JACC/delewi2013}, \emph{fragmented QRS (fQRS)} \citep{journal/Circulation/das2006}, and \emph{poor R wave progression (PRWP)} \citep{journal/IJC/kurisu2015}.
An example of each QRS abnormality is illustrated in \Leireffig{fig:method:abnormalQRS_MI_example}.

The QRS duration is the time interval between the beginning of the Q wave and the end of the S wave.
Pathological Q waves are described as the presence of Q wave with duration $\geq$ 0.03~s and/ or amplitude $\geq$ 25\% of R-wave amplitude \citep{journal/JACC/delewi2013}.
fQRS is defined as the number of additional spikes within the QRS complex \citep{journal/Circulation/das2006}.
PRWP refers to the absence of the normal increase in amplitude of the R wave in the precordial leads when advancing from lead V1 to V6 \citep{journal/IJC/kurisu2015}.
In the literature, different definitions of PRWP exist \citep{journal/JACC/mackenzie2005}.
In this work, we employ criteria including R wave amplitude of 2~mm or less in the lead V3/ V4 and the presence of reversed R-wave progression such as R of V5 $<$ R of V6 or R of V2 $<$ R of V1, or any combination of these. 

\section{Experiments and Results}

\subsection{Data Acquisition and Activation Property Configuration}
We collect 49 subjects with paired ECGs and CMR images, including cine short-axis, two- and four-chamber long-axis, localizer, and scout slices, from the UK Biobank study \citep{journal/Nature/bycroft2018}.
The locations of root nodes are set to seven fixed homologous locations to allow comparisons \citep{journal/EP/cardone2016}.
Specifically, four left ventricular (LV) earliest activation sites (LV mid-septum, LV basal-anterior paraseptal, and two LV mid-posterior) and three in the right ventricle (RV), namely, RV mid-septum and two RV free wall, are selected as root nodes. 
The CVs along the fiber, sheet, sheet-normal, and sparse/ dense endocardial directions are set to 65 cm/s, 48 cm/s, 51 cm/s, and 100/ 150 cm/s, respectively, in agreement with velocities reported for human healthy ventricular myocardium in \citep{journal/CR/myerburg1978,journal/JMCC/taggart2000}.


\subsection{Results}

 \begin{figure*}[!t] \center
    \subfigure[] {\includegraphics[width=0.52\textwidth]{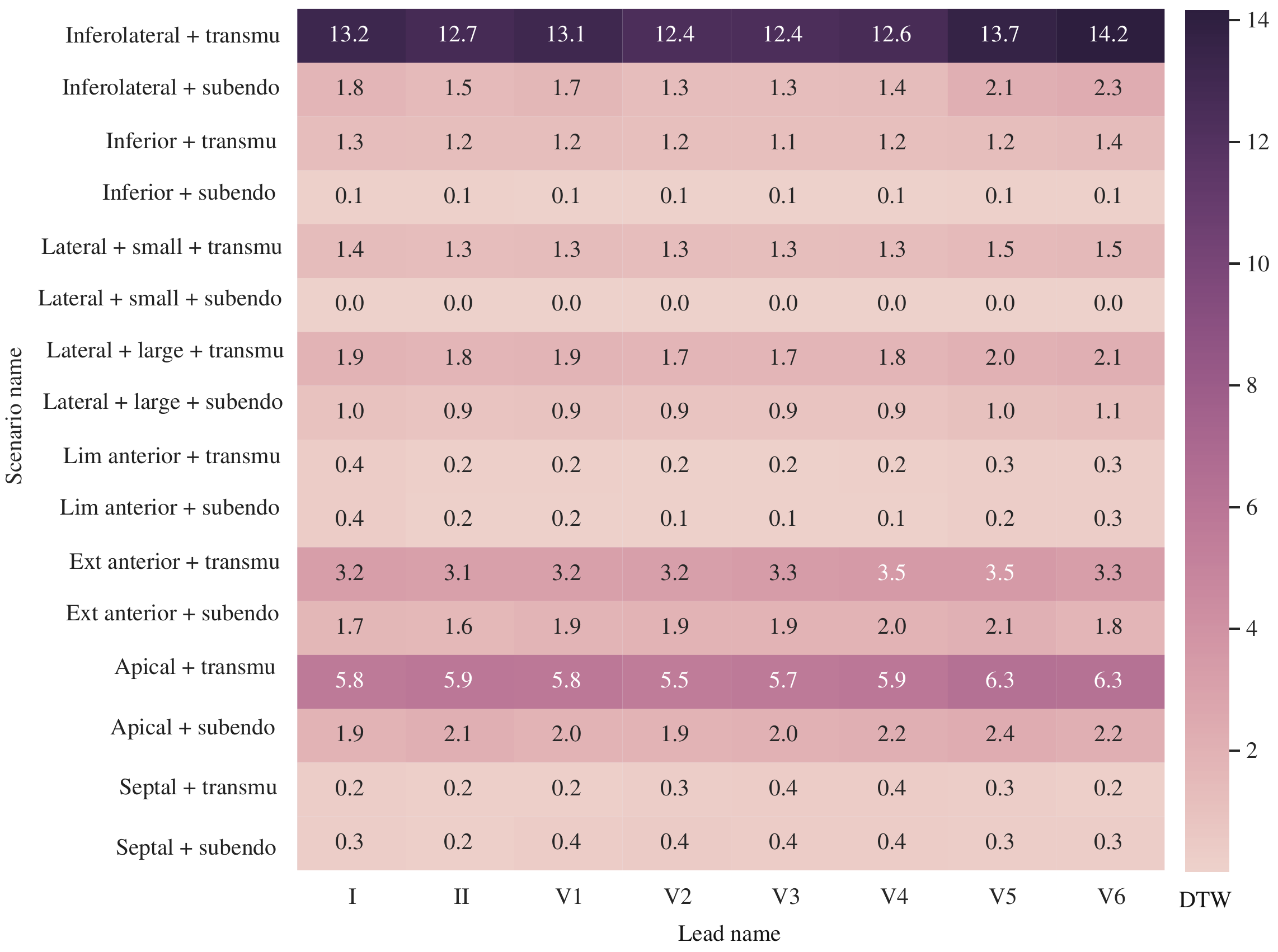}} 
    \subfigure[] {\includegraphics[width=0.471\textwidth]{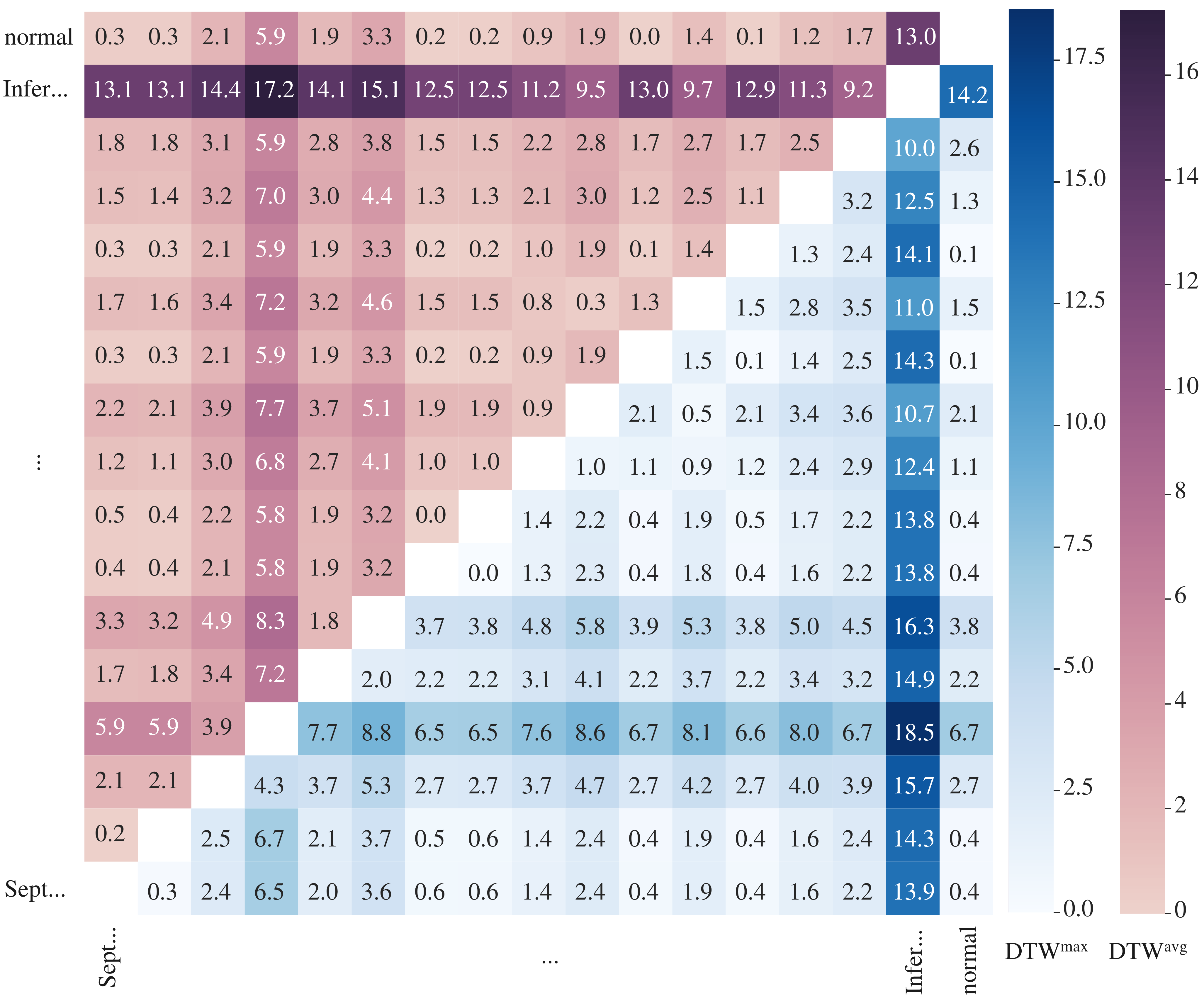}}
	\caption{
		(a) QRS dissimilarity of each MI scenario in each lead compared to the baseline; 
		(b) QRS dissimilarity between each MI scenario. The full name of MI scenario is omitted here.
        ${DTW}^{max}$ and ${DTW}^{avg}$ refer to the maximum and average dynamic time warping (DTW) values of all leads, respectively.
        transmu: tranmural; subendo: subendocardial.
		} 
	\label{fig:exp:QRS_dissimilarity}
\end{figure*}

\subsubsection{QRS Differences Depending on MI Characteristics} 
\begin{figure*}[!t] \center
    \subfigure[] {\includegraphics[width=0.49\textwidth]{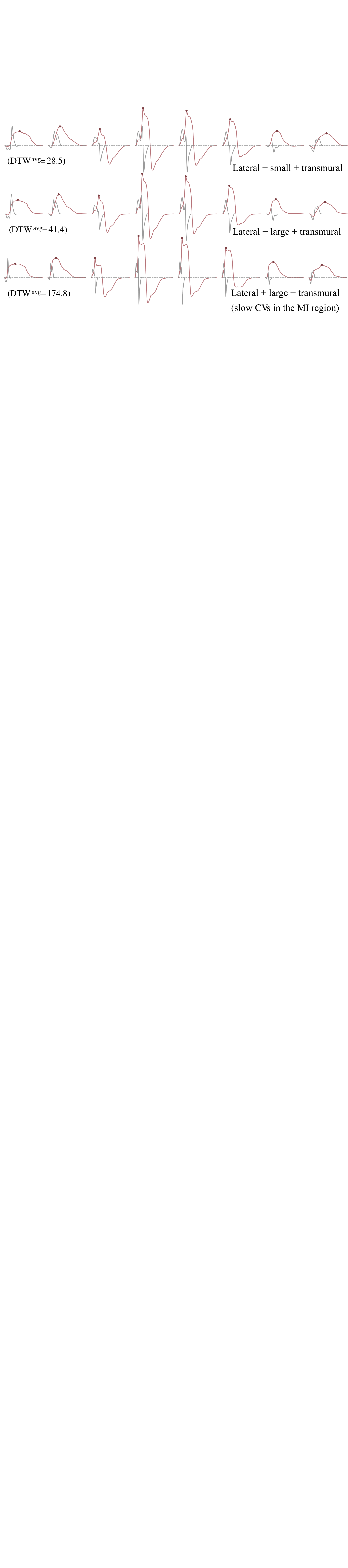}} 
    \subfigure[] {\includegraphics[width=0.49\textwidth]{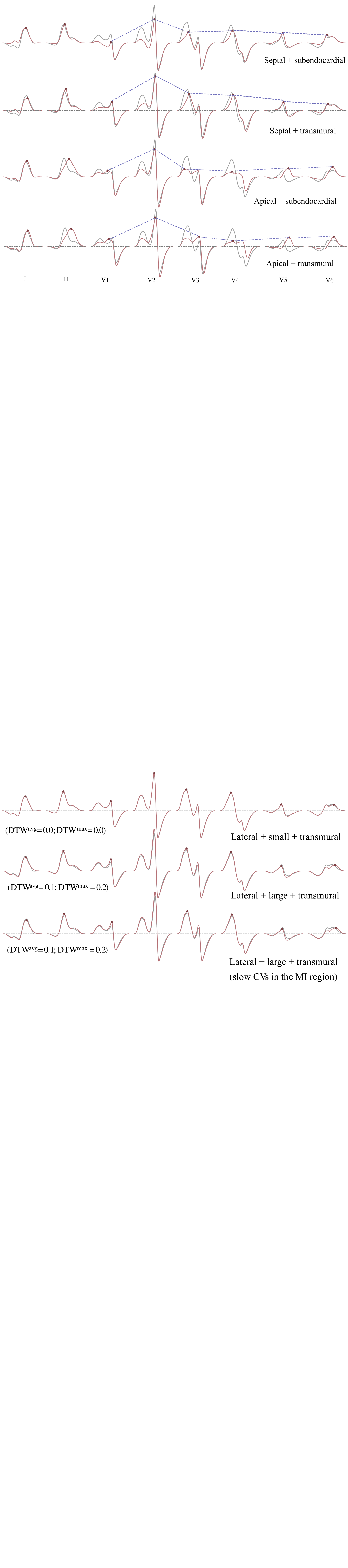}}
    \subfigure[] {\includegraphics[width=0.49\textwidth]{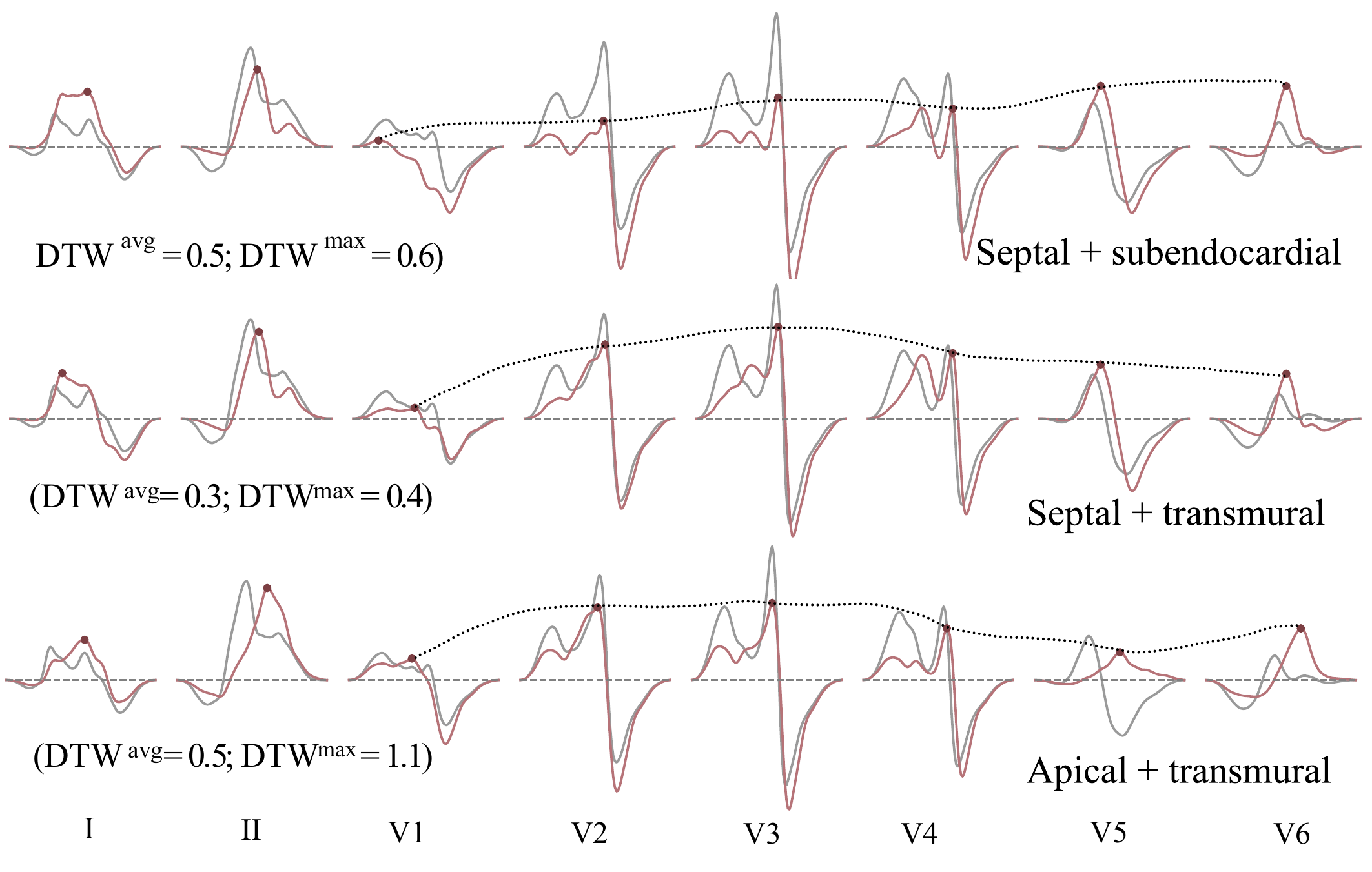}} 
    \subfigure[] {\includegraphics[width=0.49\textwidth]{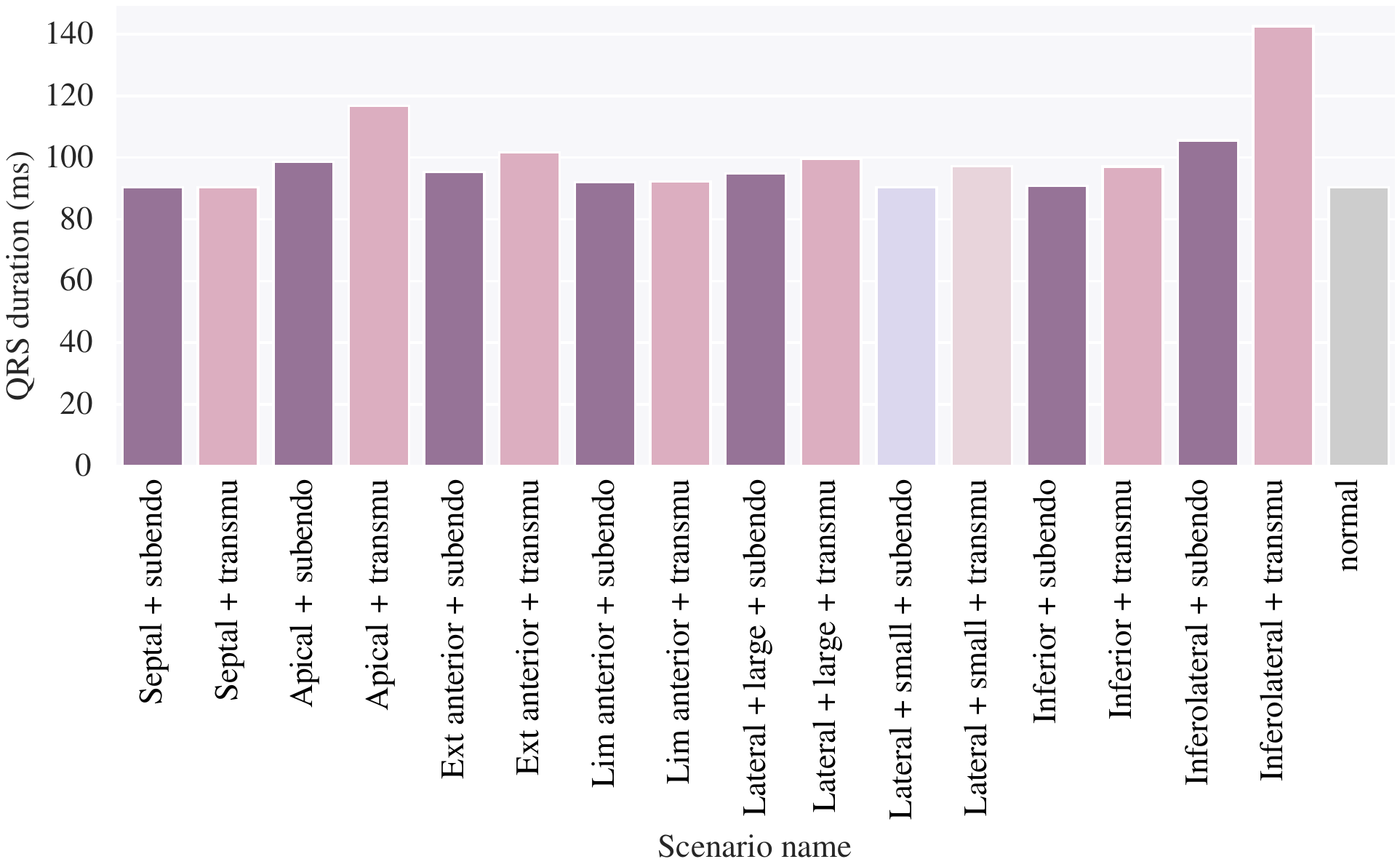}}
	\caption{
		(a-b) QRS morphology examples of lateral MIs with different sizes and CV setting. Here, MI and baseline QRS are labeled in red and grey, respectively;
        (c) QRS morphology difference among transmural and subendocardial septal MIs and poor R wave progression examples occurs in apical and septal MIs. The R wave progression is labeled with a black dashed line;
		(d) QRS duration of MI and baseline. 
		}
	\label{fig:exp:simulated_QRS_examples}
\end{figure*}

To investigate the sensitivity of QRS on the 17 MI scenarios, we compare the dissimilarity of each of these with the baseline as well as the dissimilarity between them, as shown in \Leireffig{fig:exp:QRS_dissimilarity}.
It is clear that there exist significant morphological changes in the post-MI QRS compared to the normal QRS, especially for inferolateral, extensive anterior, and apical MIs.
However, differences from healthy QRS are highly reduced, as expected, when we reduce the size of the lateral MI or its transmurality. 
In addition, there is a significant variation in the QRS of lateral MI among different subjects. 
As \Leireffig{fig:exp:simulated_QRS_examples} (a-b) shows, the QRS of lateral MI can range from substantially different to almost identical to baseline.
The extent of transmurality has evident effects on QRS morphology at each infarct location: as expected, transmural scars tend to present more evident morphological changes in the QRS than subendocardial ones.
Even for the septal scars, in which transmural and subendocardial QRS dissimilarities are the smallest (${DTW}^{max}=0.2$ and ${DTW}^{avg}=0.3$), one still can observe their morphology difference (see \Leireffig{fig:exp:simulated_QRS_examples} (c)).
Nevertheless, differences in QRS between infarct locations appear to be larger than those depending on the extent of transmurality,  suggesting that the QRS has higher sensitivity for localizing MI than predicting its transmural extent.
The major QRS morphological variation for different degrees of CV reduction setting appears to be the QRS duration, which is not unexpected.
However, according to our limited test for this purpose, we get particularly unusual QRS simulation results when we significantly reduce the CVs in the MI regions.
Therefore, the CV configuration of MI areas during simulation is still an open question that demands more exploration in the future.

The sensitivity of different QRS leads for detecting infarct location is varied.
As \Leireffig{fig:exp:QRS_dissimilarity} (a) shows, most infarct locations are represented on the QRS by leads I, V5, and V6, whereas septal MI is represented by leads V1-V4 and V3-V4 for subendocardial and transmural ones, respectively.
This result is generally consistent with those reported in clinical practice \citep{journal/CCR/nikus2014}.
In general, larger scars tend to result in QRS changes appearing in more leads.

\subsubsection{Sensitivity of Different QRS Criteria for MI Classification} 
The changes in QRS morphology for different MI scenarios are reflected in various perspectives.
Here, we introduce several QRS criteria and compare the contribution of each of these for infarct detection.
Apical, extensive anterior, and inferolateral MI tend to present prolongation of the QRS duration, as \Leireffig{fig:exp:simulated_QRS_examples} (d) shows.
PRWP mainly occurred in extensive anterior, septal, and apical MIs, similar as reported in the literature \citep{journal/IJC/kurisu2015,journal/IJC/mittal1986}. 
Specifically, the R wave amplitude in the septal MI is sometimes flattened, while the R wave of V6 tends to be larger than the R of V5 in the apical MI, as \Leireffig{fig:exp:simulated_QRS_examples} (c) shows.
The prevalence of fQRS IS more common in the inferior lead (lead II) compared with the anterior leads (leads V3 and V4) and the lateral leads (leads V5 and V6), similar to the results reported in Liu et~al. \citep{journal/JIMR/liu2020}.
The presence of fQRS in lead II and leads V3-V4 indicate inferolateral MI and extensive anterior MI, respectively.
In contrast, pathological Q wave fails to classify MI from healthy subjects in our simulation system.



\section{Discussion and Conclusion}
In this paper, we have presented a sensitivity analysis of QRS for the identification of 17 MI scenarios via Eikonal simulation.
The results have demonstrated the potential of the QRS to improve ECG-based prediction of MI characteristics and further facilitate patient-specific clinical decision-making. 
It also demonstrates the feasibility of developing a cardiac ``digital twin" deep computational model for the inference of MI.
Limitations of our study at this point include the assumption of a known set of RNs and limited variation in our anisotropic CVs.
Moreover, currently we only consider cardiac anatomical information \blue{and electrode nodes, but ignore the torso geometry}. 
Its introduction might provide relevant information about its influence in ECG patterns.
In the future, we will extend this work by introducing non-invasive personalization of the ventricular activation sequences for a more realistic representation of the cardiac conduction system.
Furthermore, this analysis could be applied on the whole ECG signal instead of only QRS, necessitating large computational costs. 
The results can be further validated from relevant clinical outcomes in the ECGs of real MI patients.
Consequently, the developed models and techniques will enable further research in cardiac digital twins.

\subsubsection{Acknowledgement.}
This research has been conducted using the UK Biobank Resource under Application Number ‘40161’. The authors express no conflict of interest.
This work was funded by the CompBioMed 2 Centre of Excellence in Computational Biomedicine (European Commission Horizon 2020 research and innovation programme, grant agreement No. 823712).
L. Li was partially supported by the SJTU 2021 Outstanding Doctoral Graduate Development Scholarship.
A. Banerjee is a Royal Society University Research Fellow and is supported by the Royal Society Grant No. URF{\textbackslash}R1{\textbackslash}221314.
The work of A. Banerjee and V. Grau was partially supported by the British Heart Foundation (BHF) Project under Grant PG/20/21/35082.

\bibliographystyle{splncs04}
\bibliography{A_refs}

\end{document}